\documentclass[runningheads]{llncs}
\usepackage{graphicx}

\usepackage{tikz}
\usepackage{comment} 
\usepackage{amsmath,amssymb} 
\usepackage{color}

\usepackage{multirow}
\usepackage{makecell}
\usepackage{subcaption}
\usepackage{bm}
\usepackage{booktabs}
\usepackage{adjustbox}

\begin{document}
\pagestyle{headings}
\mainmatter
\def\ECCVSubNumber{607}  

\title{AttentionNAS: Spatiotemporal Attention Cell Search for Video Classification} 

\titlerunning{AttentionNAS}
%
\author{
Xiaofang Wang\inst{2}\thanks{Work done while an intern at Google.} \and
Xuehan Xiong\inst{1} \and
Maxim Neumann\inst{1} \and
AJ Piergiovanni\inst{1} \and
Michael S. Ryoo\inst{1} \and
Anelia Angelova\inst{1} \and
Kris M. Kitani\inst{2} \and
Wei Hua\inst{1}
}
\authorrunning{X. Wang et al.}
%
\institute{Google \and Carnegie Mellon University}
\maketitle

\begin{abstract}
Convolutional operations have two limitations: (1) do not explicitly model where to focus as the same filter is applied to all the positions, and (2) are unsuitable for modeling long-range dependencies as they only operate on a small neighborhood. While both limitations can be alleviated by attention operations,  many design choices remain to be determined to use attention, especially when applying attention to videos. Towards a principled way of applying attention to videos, we address the task of spatiotemporal attention cell search. We propose a novel search space for spatiotemporal attention cells, which allows the search algorithm to flexibly explore various design choices in the cell. The discovered attention cells can be seamlessly inserted into existing backbone networks, e.g., I3D or S3D, and improve video classification accuracy by more than 2\% on both Kinetics-600 and MiT datasets. The discovered attention cells outperform non-local blocks on both datasets, and demonstrate strong generalization across different modalities, backbones, and datasets. Inserting our attention cells into I3D-R50 yields state-of-the-art performance on both datasets.

\end{abstract}

\section{Introduction}

One major contributing factor to the success of neural networks in computer vision is the novel design of network architectures. In early work, most network architectures~\cite{krizhevsky2012imagenet,szegedy2015going,he2016deep} were manually designed by human experts based on their knowledge and intuition of specific tasks. Recent work on neural architecture search (NAS)~\cite{zoph2016neural,zoph2018learning,liu2018darts,liu2018progressive,real2019regularized} proposes to directly learn the architecture for a specific task from data and discovered architectures have been shown to outperform human-designed ones.

Convolutional Neural Networks (CNNs) have been the \emph{de facto} architecture choice. Most work in computer vision uses convolutional operations as the primary building block to construct the network. However, convolutional operations still have their limitations. It has been shown that attention is complementary to convolutional operations, and they can be combined to further improve performance on vision tasks~\cite{woo2018cbam,wang2018non,bello2019attention}.

While being complementary to convolution, many design choices remain to be determined to use attention. The design becomes more complex when applying attention to videos, where the following questions arise: \textit{What is the right dimension to apply an attention operation to videos? Should an operation be applied to the temporal, spatial, or spatiotemporal dimension? How to compose multiple attention operations applied to different dimensions?}

Towards a principled way of applying attention to videos, we address the task of spatiotemporal attention cell search, i.e., the automatic discovery of cells that use attention operations as the primary building block. The discovered attention cells can be seamlessly inserted into a wide range of backbone networks, e.g., I3D~\cite{carreira2017quo} or S3D~\cite{xie2018rethinking}, to improve the performance on video understanding tasks.

Specifically, we propose a search space for spatiotemporal attention cells, which allows the search algorithm to flexibly explore all of the aforementioned design choices in the cell. The attention cell is constrcuted by composing several primitive attention operations. Importantly, we consider two types of primitive attention operations: (1) map-based attention~\cite{Park2018BAMBA,woo2018cbam} and (2) dot-product attention (a.k.a., self-attention)~\cite{vaswani2017attention,wang2018non,bello2019attention}. 
Map-based attention explicitly models where to focus in videos, compensating for the fact that convolutional operations apply the same filter to all the positions in videos. Dot-product attention enables the explicit modeling of long-range dependencies between distant positions in videos, accommodating the fact that convolutional operations only operate on a small and local neighborhood.

We aim to find an attention cell from the proposed search space such that the video classification accuracy is maximized when adding that attention cell into the backbone network. But the search process can be extremely costly. One significant bottleneck of the search is the need to constantly evaluate different attention cells. Evaluating the performance of an attention cell typically requires training the selected attention cell as well as the backbone network from scratch, which can take days on large-scale video datasets, e.g., Kinetics-600~\cite{carreira2018short}.

To alleviate this bottleneck, we consider two search algorithms: (1) Gaussian Process Bandit (GPB)~\cite{Srinivas2009GaussianPO,snoek2012practical}, which judiciously selects the next attention cell for evaluation based on the attention cells having been evaluated so far, allowing us to find high-performing attention cells within a limited number of trials; (2) differentiable architecture search~\cite{liu2018darts}, where we develop a differentiable formulation of the proposed search space, making it possible to jointly learn the attention cell design and network weights through back-propagation, without explicitly sampling and evaluating different cells. The entire differentiable search process only consumes a computational cost similar to fully training one network on the training videos. This formulation also allows us to learn position-specific attention cell designs with zero extra computational cost (see Sec~\ref{sec-differentiable} for details).

We conduct extensive experiments on two benchmark datasets: Kinetics-600~\cite{carreira2018short} and Moments in Time (MiT)~\cite{monfort2019moments}. Our discovered attention cells can improve the performance of two backbone networks I3D~\cite{carreira2017quo} and S3D~\cite{xie2018rethinking} by more than 2\% on both datasets, and also outperforms non-local blocks -- the state-of-the-art manually designed attention cells for videos. Inserting our attention cells into I3D-R50~\cite{wang2018non} yields state-of-the-art performance on both datasets. Notably, our discovered attention cells can also generalize well across modalities (RGB to optical flow), backbones (e.g., I3D to S3D or I3D to I3D-R50), and datasets (MiT to Kinetics-600 or Kinetics-600 to MiT).

\textbf{Contributions:}
(1) This is the first attempt to extend NAS beyond discovering convolutional cells to attention cells. (2) We propose a novel search space for spatiotemporal attention cells that use attention operations as the primary building block, which can be seamlessly inserted into existing backbone networks to improve their performance on video classification. (3) We develop a differentiable formulation of the proposed search space, making it possible to learn the attention cell design with back-propagation and learn position-specific attention cell designs with zero extra cost. (4) Our discovered attention cells outperform non-local blocks, on both the Kinetics-600 and MiT dataset. We achieve state-of-the-art performance on both datasets by inserting our discovered attention cells into I3D-R50. Our attention cells also demonstrate strong generalization capability when being applied to different modalities, backbones, or datasets.

\begin{figure}[t]
\centering
\includegraphics[width=0.85\columnwidth]{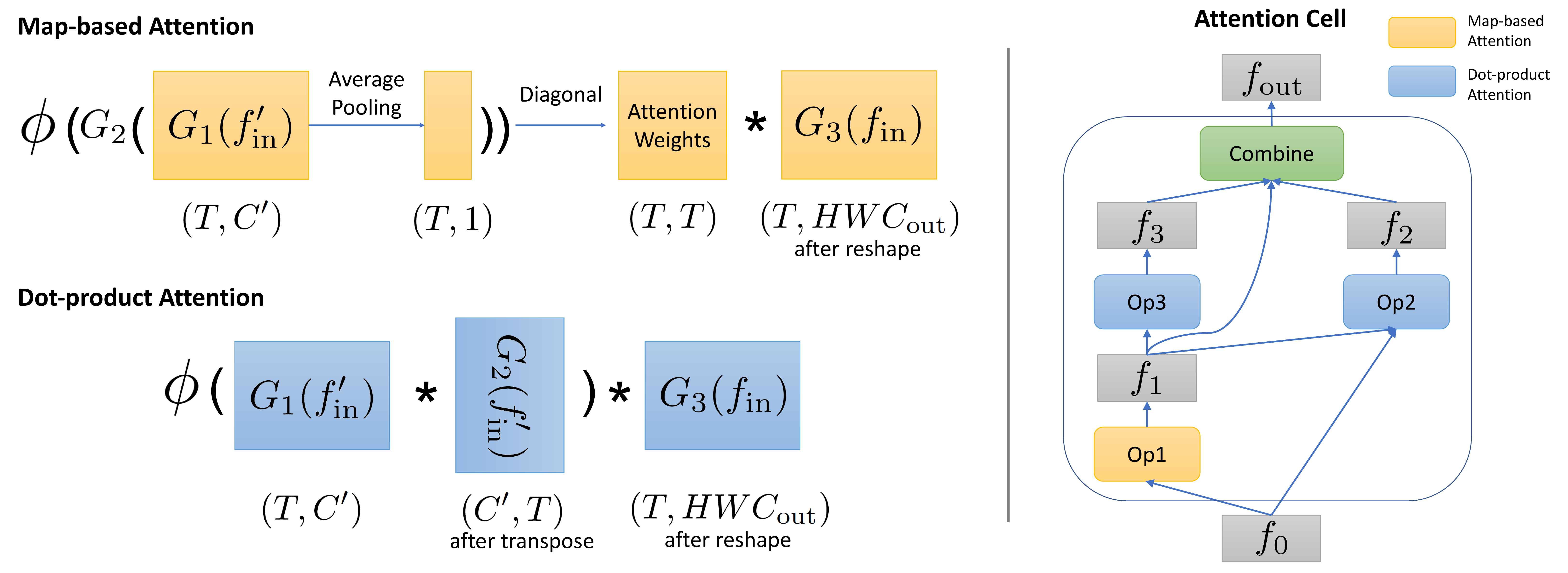}
\caption{Illustration of the operation-level search space (left) and cell-level search space (right). The example attention operations use temporal as the attention dimension and the tuple under each feature map denotes its shape.}
\label{fig-search-space}
\end{figure}

\section{Related Work}

\textbf{Video Classification.} Early work on video classification extends image classification CNNs with recurrent networks~\cite{donahue2015long,yue2015beyond} or two-stream architectures~\cite{simonyan2014two,feichtenhofer2016convolutional} that take both RGB frames and optical flow frames as inputs. Recent work on video classification are mainly based on 3D convolution~\cite{tran2015learning} or its variants to directly learn video representations from RGB frames. I3D~\cite{carreira2017quo} proposes to inflate the filters and pooling kernels of a 2D CNN into 3D to leverage successful 2D CNN architecture designs and their ImageNet pretrained weights. S3D~\cite{xie2018rethinking} improves upon I3D by decomposing a 3D convolution into a 2D spatial convolution and a 1D temporal convolution. A similar idea is also explored in P3D~\cite{qiu2017learning}. CPNet~\cite{liu2019learningvideo} learns video representations by aggregating information from potential correspondences. SlowFast~\cite{feichtenhofer2019slowfast} proposes an architecture operating at two different frame rates, where spatial semantics are learned on low frame rates, and temporal dynamics are learned on high frame rates. Different from them, we do not focus on proposing novel CNN architecture designs for video classification. Instead, we focus on discovering attention cells using attention operations as the primary building block, which are complementary to CNNs.

\textbf{Attention in Vision.}  Both map-based attention and dot-product attention are useful for computer vision tasks. Map-based attention~\cite{Park2018BAMBA,woo2018cbam} has been used to improve the performance of CNNs on image recognition, where spatial attention maps are learned to scale the features given by convolutional layers. Dot-product attention~\cite{vaswani2017attention} is successfully used in sequence modeling and transduction tasks, e.g., machine translation, and is recently used to augment CNNs and enhances their performance on image recognition~\cite{bello2019attention}. Non-local blocks~\cite{wang2018non} are proposed to capture long-range dependencies in videos and can significantly improve the video classification accuracy of CNNs. Non-local blocks can be viewed as applying one single dot-product attention operation to the spatiotemporal dimension. In contrast, our attention cells can contain multiple attention operations applied to different dimensions of videos. Non-local blocks are a particular case in our proposed search space, and our attention cells are discovered automatically in a data-driven way instead of being manually designed.

\textbf{NAS - Search Space.} Search space is crucial for NAS. Randwire~\cite{xie2019exploring} shows that one random architecture from a carefully designed search space can achieve competitive performance on image recognition. NASNet~\cite{zoph2018learning} proposes to search for convolutional cells that can be stacked multiple times to form the entire architecture. Auto-DeepLab~\cite{liu2019auto} proposes a two-level hierarchical architecture search space for semantic image segmentation. AssembleNet~\cite{Ryoo2020AssembleNet:} proposes to search for the connectivity between multi-stream convolutional blocks for video classification. They all focus on finding convolutional cells or networks for the end task. Different from them, our proposed search space uses attention as the primary building component instead of convolution.

\textbf{NAS - Search Algorithm.} Various search algorithms have been explored in NAS, such as random search~\cite{li2019random,Yu2020Evaluating}, reinforcement learning~\cite{baker2016designing,zoph2016neural,zoph2018learning,zhong2018practical}, evolutionary algorithms~\cite{xie2017genetic,real2017large,real2019regularized}, Bayesian optimization (BO)~\cite{kandasamy2018neural,cao2018learnable}, and differentiable methods~\cite{liu2018darts}. We have tried using GPB (belonging to the category of BO) to search for desired attention cells. We also develop a differentiable formulation of our proposed search space. This makes it possible to conduct the search using differentiable methods and greatly improves the search speed.

\section{Attention Cell Search Space}
\label{sec-searchspace}

We aim to search for spatiotemporal attention cells, which can be seamlessly inserted into a wide range of backbone networks, e.g., I3D~\cite{carreira2017quo} or S3D~\cite{xie2018rethinking}, to improve the performance on video understanding tasks.

Formally, an attention cell takes a 4D feature map of shape $(T, H, W, C)$ as input and outputs a feature map of the same shape. $T, H$, and $W$ are the temporal dimension, height, and width of the feature map, respectively. $C$ denotes the number of channels. The output of an attention cell is enforced to have the same shape as its input by design, so that the discovered attention cells can be easily inserted after any layers in any existing backbone networks.

An attention cell is composed of $K$ primitive attention operations. The proposed attention cell search space consists of an operation level search space and a cell level search space (see Fig.~\ref{fig-search-space}). The operation level search space contains different choices to instantiate an individual attention operation. The cell level search space consists of different choices to compose the $K$ operations to form a cell, i.e., the connectivity between the $K$ operations within a cell. We first introduce the operation level search space and then the cell level search space.

\subsection{Operation Level Search Space}

An attention operation takes a feature map of shape $(T, H, W, C_{\text{in}})$ as input and outputs an attended featured map of shape $(T, H, W, C_{\text{out}})$. For an attention operation, $C_{\text{in}}$ and $C_{\text{out}}$ can be different. To construct an attention operation, we need to make two fundamental choices: the dimension to compute the attention weights and the type of the attention operation.

\subsubsection{Attention Dimension}
For brevity, we term the dimension to compute the attention weights as \textit{attention dimension}. In CNNs for video classification, previous work~\cite{qiu2017learning,xie2018rethinking,feichtenhofer2019slowfast} has studied when to use temporal convolution (e.g., $3 \times 1 \times 1$), spatial convolution (e.g., $1 \times 3 \times 3$), and spatiotemporal convolution  (e.g., $3 \times 3 \times 3$). It is also a valid question to ask for attention what is the right dimension to apply an attention operation to videos: temporal, spatial or spatiotemporal (temporal and spatial together). The choice of the attention dimension is important as computing attention weights for different dimensions represents focusing on different aspects of the video.

\subsubsection{Attention Operation Type}
We consider two types of attention operations, each of which helps address a specific limitation of convolutional operations, as mentioned in the introduction:
\begin{itemize}
\item \textbf{Map-based attention}~\cite{Park2018BAMBA,woo2018cbam}: Map-based attention learns a weighting factor for each position in the attention dimension and scales the feature map with the learned attention weights. Map-based attention explicitly models what positions in the attention dimension to attend to in videos.
\item \textbf{Dot-product attention}~\cite{vaswani2017attention,wang2018non,bello2019attention}: A dot-product attention operation computes the feature response at a position as a weighted sum of features of all the positions in the attention dimension, where the weights are determined by a similarity function between features of all the positions~\cite{wang2018non,bello2019attention}. Dot-product attention explicitly models the long-range interactions among distant positions in the attention dimension.
\end{itemize}

We now describe the details of the two types of attention operations. Let $f_\text{in}$ denote the input feature map to an attention operation and denote its shape as $(T, H, W, C_{\text{in}})$. Applying an attention operation consists of three steps, including reshaping the input feature map $f_\text{in}$, computing the attention weights, and applying the attention weights.

\paragraph{Reshape $f_\text{in}$.} We reshape $f_\text{in}$ into a 2D feature map $f'_\text{in}$ before computing the attention weights. The first dimension of $f'_\text{in}$ is the attention dimension and the second dimension contains the remaining dimensions. For example, $f'_\text{in}$ has the shape of $(T, HWC_{\text{in}})$ when temporal is the attention dimension and has the shape of $(THW, C_{\text{in}})$ when spatiotemporal is the attention dimension. We denote this procedure as a function $\texttt{ReshapeTo2D}$, i.e., $f'_\text{in} = \texttt{ReshapeTo2D}(f_\text{in})$.

Spatial attention requires extra handling. As video content changes over time, when applying attention to the spatial dimension, each frame $f_\text{in}^t$ should have its own spatial attention weights, where $f_\text{in}^t$ is the $t^{th}$ frame in $f_\text{in}$ and has the shape of $(H, W, C_\text{in})$. Therefore, when spatial is the attention dimension, instead of reshaping the entire 4D feature map $f_\text{in}$, we reshape $f_\text{in}^t$ into a 2D feature map $f_\text{in}^{\prime t}$ of shape $(HW, C_\text{in})$ for every $t$, i.e., $f_\text{in}^{\prime t}=\texttt{ReshapeTo2D}(f_\text{in}^t) (1 \leq t \leq T)$.

\paragraph{Map-based attention.} Assuming temporal is the attention dimension, map-based attention generates $T$ attention weights to scale the feature map of each temporal frame. The attention weights are computed as follows:
\begin{equation}
\label{eq-mapatt}
W_\text{map} = \texttt{Diag}(\phi(G_2(\texttt{AvgPool}(G_1(f'_\text{in}))))).
\end{equation}
$G_1$ is a 1D convolutional layer with kernel size as 1, which reduces the dimension of the feature response of each temporal frame from $HWC_{\text{in}}$ to $C'$ and gives a feature map of shape $(T, C')$. $\texttt{AvgPool}$ denotes an average pooling operation applied to each temporal dimension and outputs a $T$-dim vector.  The multilayer perceptron $G_2$ and the activation function $\phi$ (e.g., the sigmoid function) further transform the $T$-dim vector to $T$ attention weights. More details about the activation function are discussed later. $\texttt{Diag}$ rearranges the $T$ attention weights into a $T \times T$ matrix, where the $T$ attention weights are placed on the diagonal of the matrix. The obtained attention weight matrix $W$ is a diagonal matrix.

Similarly, when spatiotemporal is the attention dimension, map-based attention gives a $THW \times THW$ diagonal matrix containing the attention weights. When spatial is the attention dimension, we generate one $HW \times HW$ diagonal matrix for every $f_{\text{in}}^{\prime t}$ ($1\leq t \leq T$) separately, using the above described procedure. Note that while different frames have separate spatial attention weights, $G_1$ and $G_2$ are shared among different frames when computing attention weights.

\paragraph{Dot-product attention.} When applying dot-product attention to the temporal dimension, a $T\times T$ attention weight matrix is generated as follows:
\begin{equation}
\label{eq-dotproductatt}
W_\text{dot-prod} = \phi(G_1(f'_\text{in})  G_2(f'_\text{in})^{T}).
\end{equation}
Here, $G_1$ and $G_2$ are both a 1D convolutional layer with kernel size as 1 and they both output a feature map of shape $(T, C')$. Let $Q = G_1(f'_\text{in})$ and $K = G_2(f'_\text{in})$. $QK^{T}$ computes an similarity matrix between the features of all the temporal frames. We then use $\phi$, an activation function of our choice, e.g., the softmax function, to convert the similarity matrix into attention weights. Note that different from $W_\text{map}$, $W_\text{dot-prod}$ is a full matrix instead of a diagonal matrix.

When being applied to the spatiotemporal dimension, dot-product attention generates a $THW \times THW$ attention weight matrix. When applying dot-product attention to the spatial dimension, each frame has its own attention weights (a $HW \times HW$ matrix), where  $G_1$ and $G_2$ are shared among different frames.

\paragraph{Apply the attention weights.} We apply the attention weight matrix to the input feature map through matrix multiplication to obtain the attended feature map:
\begin{equation}
\label{eq-applyattweights}
\begin{split}
f_\text{out}  = \texttt{ReshapeTo2D}^{-1} (  W \texttt{ReshapeTo2D}(G_3(f_\text{in})) ).
\end{split}
\end{equation}
$W$ is the weight matrix generated by map-based attention $(W_\text{map})$ or dot-product attention $(W_\text{dot-prod})$. $G_3$ is a $1\times 1 \times 1$ convolutional layer to reduce the number of channels of $f_\text{in}$ from $C_\text{in}$ to $C_\text{out}$. If temporal is the attention dimension, $W$ has the shape of $(T, T)$ and $\texttt{ReshapeTo2D}(G_3(f_\text{in}))$ has the shape $(T, HWC_\text{out})$. $\texttt{ReshapeTo2D}^{-1}$ is the inverse function of $\texttt{ReshapeTo2D}$, reshaping the attended feature map back to the shape of $(T, H, W, C_\text{out})$.

For spatial attention, the attention weights are applied to each frame independently, i.e., $f^t_\text{out}  = \texttt{ReshapeTo2D}^{-1} (  W^t \texttt{ReshapeTo2D}(G_3(f^t_\text{in})) )$, where $W^t$ is the spatial attention weights for frame $t$ and $f^t_\text{out}$ has the shape of $(H, W, C_\text{out})$. We stack $\{f^t_\text{out}\mid 1\leq t \leq T\}$ along the temporal dimension to form the attended feature map $f_\text{out}$ of shape $(T, H, W, C_\text{out})$. Similar to $G_1$ and $G_2$ used for computing attention weights, $G_3$ is also shared among different frames.

Note that by design $G_3$ only changes number of channels, i.e., transforms the features at each spatiotemporal position. The spatiotemporal structure of the input $f_\text{in}$ is preserved. This ensures that after the application of attention weights, $f_\text{out}$ still follows the original spatiotemporal structure of the input $f_\text{in}$.

\paragraph{Activation function.} We empirically find that the activation function $\phi$ (see Eq.~\ref{eq-mapatt} and Eq.~\ref{eq-dotproductatt}) used in the attention operation can influence the performance. So, we also include the choice of the activation function in the operation level search space and rely on the search algorithm to choose the right one for each attention operation. We consider the following four choices for the activation function: (1) no activation function, (2) ReLU, (3) sigmoid, and (4) softmax.

\subsection{Cell Level Search Space}
We define an attention cell as a cell composed of $K$ attention operations. Let $f_0$ denote the input feature map to the entire attention cell and $(T, H, W, C)$  be the shape of $f_0$. $f_0$ is usually the output of a stack of convolutional layers. An attention cell takes $f_0$ as input and outputs a feature map of the same shape.

The connectivity between convolutional layers is essential to the performance of CNNs, no matter if the network is manually designed, e.g., ResNet~\cite{he2016deep} and Inception~\cite{szegedy2015going}, or automatically discovered~\cite{zoph2016neural,zoph2018learning,xie2019exploring}. Similarly, to build an attention cell, another critical design choice is how the $K$ attention operations are connected inside the cell, apart from the design of these attention operations.

As shown in Fig.~\ref{fig-search-space}, in an attention cell, the first attention operation always takes $f_0$ as input and outputs feature map $f_1$. The $k^{th} (2 \leq k \leq K)$ attention operation chooses its input from $\{f_0, f_1, \ldots, f_{k-1}\}$ and gives feature map $f_k$ based on the selected input. We allow the $k^{th}$ operation to choose multiple feature maps from $\{f_0, f_1, \ldots, f_{k-1}\}$ and compute a weighted sum of selected feature maps as its input, where the weights are learnable parameters. This process is repeated for all $k$ and allows us to explore all possible connectivities between the $K$ attention operations in the cell.

We combine $\{f_1, f_2, \ldots, f_K\}$ to obtain the output feature map of the entire attention cell. For all attention operations inside the cell, we set their output shape to be $(T, H, W, C_\text{op})$, i.e., $f_k$ has the shape of $(T, H, W, C_\text{op})$ for all $k (1 \leq k \leq K)$. $C_{\text{op}}$ is usually smaller than $C$ to limit the computation in an attention cell with multiple attention operations. We concatenate $\{f_1, f_2, \ldots, f_K\}$ along the channel dimension and then employ a $1\times 1 \times 1$ convolution to transform the concatenated feature map back to the same shape as the input $f_0$. We denote the feature map after transformation as $f_\text{comb}$. Similar to non-local blocks~\cite{wang2018non}, we add a residual connection between the input and output of the attention cell. So the final output of the attention cell is the sum of $f_0$ and $f_\text{comb}$. The combination procedure is the same for all attention cells.

\section{Search Algorithm}

\subsection{Gaussian Process Bandit (GPB)}
Given $K$, i.e., the number of attention operations inside the attention cell, the attention cell design can be parameterized by a fixed number of hyper-parameters, including the attention dimension, the type and the activation function of each attention operation, and the input to each attention operation.

We employ GPB~\cite{Srinivas2009GaussianPO,snoek2012practical}, a popular hyper-parameter optimization algorithm, to optimize all the hyper-parameters for the attention cell design jointly. Intuitively, GPB can predict the performance of an attention cell at a modest computational cost without actually training the entire network, based on those already evaluated attention cells. Such prediction helps GPB to select promising attention cells to evaluate in the following step and makes it possible to discover high-performing attention cells within a limited number of search steps.

Concretely, in GPB, the performance of an attention cell is modeled as a sample from a Gaussian process. At each search step, GPB selects the attention cell for evaluation by optimizing the Gaussian process upper confidence conditioned on those already evaluated attention cells.

\begin{figure}[t]
\centering
\includegraphics[width=0.85\columnwidth]{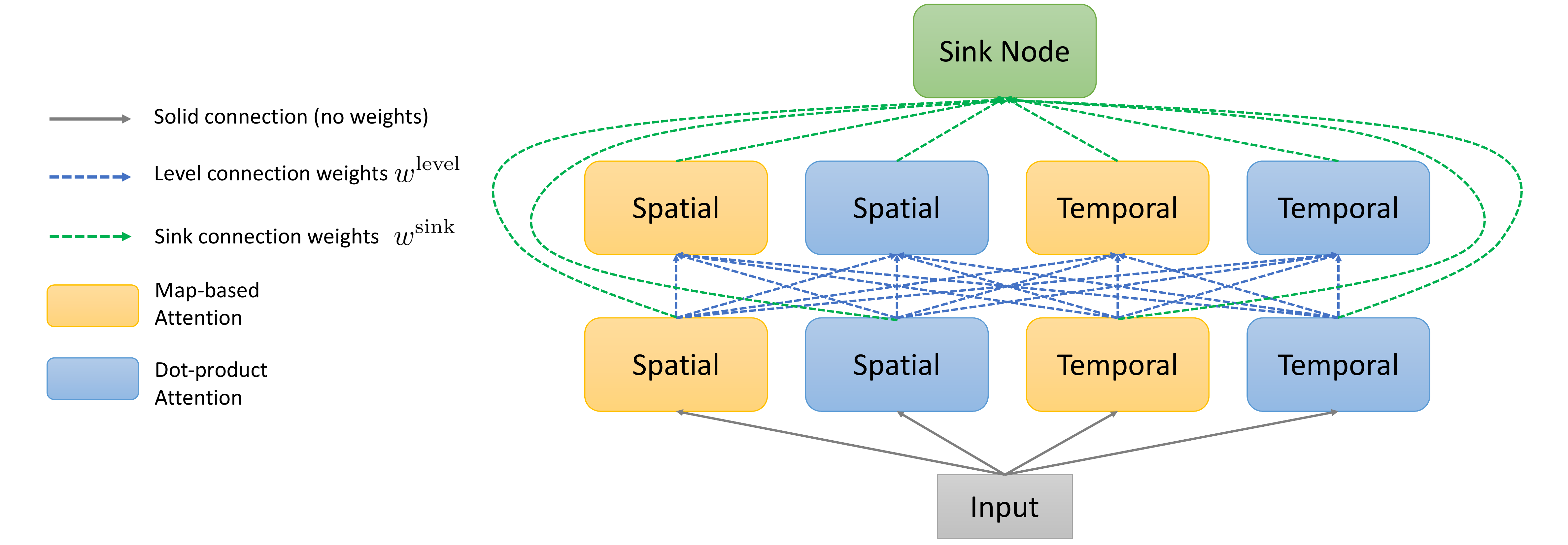}
\caption{Illustration of the supergraph used by the differentiable method. }
\label{fig-supergraph}
\end{figure}

\subsection{Differentiable Architecture Search}
\label{sec-differentiable}

Inspired by recent progress on differentiable architecture search~\cite{liu2018darts}, we develop a differentiable formulation of our proposed search space. The formulation makes it possible to jointly learn the attention cell design and network weights with back-propagation, without explicitly sampling and evaluating different cells.

\subsubsection{Differentiable Formulation of Search Space}
We propose to represent the attention cell search space as a supergraph, where all the possible attention cells are different subgraphs of this supergraph. The supergraph representation allows us to parameterize the design of an attention cell with a set of continuous and differentiable connection weights between the nodes in the supergraph.

To be more specific, we define the supergraph to have $m$ levels, where each level has $n$ nodes. Each node is an attention operation of a pre-defined type (map-based or dot-product attention) and a pre-defined attention dimension. Fig.~\ref{fig-supergraph} shows an example supergraph with 2 levels, where each level has 4 nodes. The input feature map to the entire attention cell is passed to all the nodes at the first level. Starting from the second level, the input feature map to a node is a weighted sum of the output feature maps of all the nodes at its previous level:
\begin{equation}
f_{i, j}^{\text{in}} = \sum_{k=1}^{n} w_{i, j, k}^{\text{level}} \cdot f_{i-1, k}^{\text{out}},
\end{equation}
where $2 \leq  i \leq m$, $1 \leq j \leq n$, $f_{i, j}^{\text{in}}$ is the input to the $j^{th}$ node at $i^{th}$ level, $f_{i-1, k}^{\text{out}}$ is the output of the $k^{th}$ node at $(i-1)^{th}$ level, and $w_{i, j}^{\text{level}}$ are the connection weights between the $j^{th}$ node at $i^{th}$ level and all the nodes at $(i-1)^{th}$ level. In practice, $w_{i, j}^{\text{level}}$ is a probability distribution obtained by softmax.

For each node in the supergraph, we also learn a probability distribution over the possible choices of activation functions. The output of a node is a weighted sum of the attended feature map under different activation functions:
\begin{equation}
f^{\text{out}}_{i,j} = \sum_{k=1}^{|\mathcal{A}|} w_{i, j, k}^{\text{activation}} \cdot f_{i,j}^{\text{out}, \phi_k},
\end{equation}
where $\mathcal{A}$ is the set of available activation functions, $\phi_k$ is the $k^{th}$ activation function in $\mathcal{A}$, $w_{i, j, k}^{\text{activation}}$ is the weighting factor to be learned for $\phi_k$, and  $f_{i,j}^{\text{out}, \phi_k}$ is the attended feature map under the activation function $\phi_k$. The only difference among these attended feature maps $\{f_{i,j}^{\text{out}, \phi_k}\}$ is the activation function $\phi$ used in Eq.~\ref{eq-mapatt} or Eq.~\ref{eq-dotproductatt}. The layers $G_1$, $G_2$ and $G_3$ are shared by different activation functions within one node.

The supergraph has a sink node, receiving the output feature maps of all the nodes. The sink node is defined as follows:
\begin{equation}
f_{\text{sink}}^{\text{out}} = \sum_{1\leq i \leq m, 1\leq j \leq n} w_{i, j}^{\text{sink}} \cdot G_{i,j}(f_{i, j}^{\text{out}}),
\end{equation}
where $f_{\text{sink}}^{\text{out}}$ is the output of the sink node, $f_{i, j}^{\text{out}}$ is the output of the $j^{th}$ node at $i^{th}$ level, $G_{i,j}$ is a $1 \times 1 \times 1$  convolutional layer changing the number of channels in $f_{i, j}^{\text{out}}$ to $C$, and $w_{i, j}^{\text{sink}}$ is the weighting factor to be learned. We enforce $f_{\text{sink}}^{\text{out}}$ to have the same shape as the input to the supergraph, so that the supergraph can be inserted into any position of the backbone network. Same as attention cells, a residual connection is added between the input and output of the supergraph.

\subsubsection{Attention Cell Design Learning} Both the network weights, e.g., weights of convolutional layers in the network, and the connection weights in the supergraph ($\{w^{\text{level}}, w^{\text{sink}}, w^{\text{activation}} \}$) are differentiable. 
During the search, we insert supergraphs into the backbone network and jointly optimize the network weights and connection weights by minimizing the training loss using gradient descent. The entire search process only consumes a computational cost similar to fully training one network on the training videos. Once the training is completed, we can derive the attention cell design from the learned connection weights.

Note that we insert the supergraphs at positions where the final attention cells will be inserted. In practice, usually multiple supergraphs or attention cells (e.g., 5) are inserted into the backbone network. If we enforce the inserted supergraphs to share the same set of connection weights, we will obtain one single attention cell design, dubbed as the \textit{position-agnostic} attention cell.

One significant advantage of the differentiable method is that we can also learn separate connection weights for supergraphs inserted at different positions, which will give \textit{position-specific} attention cells (see Table~\ref{table-differentiable}). Searching for separate attention cells for different positions results in an exponentially larger search space than searching for one single attention cell. But thanks to the differentiable method, we can learn position-specific attention cells with zero extra cost compared to learning one position-agnostic attention cell.

\subsubsection{Attention Cell Design Derivation} We derive the attention cell design from the learned continuous connection weights. We first choose the top $\alpha$ nodes with the highest weights in $w^{\text{sink}}$ and add them to the set $S$. Then for each node in $S$, we add its top $\beta$ predecessors in its previous level to $S$, based on the corresponding connection weights in $w^{\text{level}}$. This process is conducted recursively for every node in $S$ until we reach the first level. $\alpha$ and $\beta$ are two hyper-parameters.

Recall that each node is an attention operation of a pre-defined type and attention dimension. So,  $S$ contains a set of selected attention operations. The construction process of $S$ also determines how these attention operations are connected. For all the selected attention operations, we decide its activation function based on the corresponding weighting factors in $w^{\text{activation}}$.

\section{Experiments}

\subsection{Experimental Setup}
\textbf{Datasets.} We conduct experiments on two benchmark datasets: Kinetics-600~\cite{carreira2018short} and Moments in Time (MiT)~\cite{monfort2019moments}. Top-1 and top-5 classification accuracy are used as the evaluation metric for both datasets.

\textbf{Backbones.} We conduct the attention cell search on two backbones: I3D~\cite{carreira2017quo} and S3D~\cite{xie2018rethinking}. Both I3D and S3D are constructed based on the Inception~\cite{szegedy2015going} network. When examining the generalization of the found cells, we also consider the backbone I3D-R50~\cite{wang2018non}, which is constructed based on ResNet-50~\cite{he2016deep}.

\textbf{Baselines.} Non-local blocks~\cite{wang2018non} are the state-of-the-art manually designed attention cell for video classification and are the most direct competitor of our automatically searched attention cells. We mainly focus on the relative improvement brought by our attention cells after being inserted into backbones. Besides non-local blocks, we also compare with other state-of-the-art methods for video classification, such as TSN~\cite{wang2016temporal}, TRN~\cite{zhou2018temporal}, and SlowFast~\cite{feichtenhofer2019slowfast}.
 
\begin{table}[t]
\centering
\caption{Search results on Kinetics-600 and MiT using GPB. Our attention cells improve the classification accuracy for both backbones and on both datasets.}
\label{table-gpb}
\begin{tabular}{llccccccccccc}
\toprule
 & && & \multicolumn{3}{c}{\textbf{Kinetics}} && & \multicolumn{3}{c}{\textbf{MiT}} \\  & Model && & Top-1 & Top-5 & $\Delta$Top-1 && & Top-1 & Top-5 & $\Delta$Top-1 \\ \midrule

I3D & Backbone~\cite{carreira2017quo} && & 75.58 & 92.93 & - && & 27.38 & 54.29 & - \\
 & Non-local~\cite{wang2018non} && & 76.87 & 93.44 & 1.29 && & \textbf{28.54} & 55.35 & \textbf{1.16} \\
 & Ours - GPB && & \textbf{77.39} & \textbf{93.63} & \textbf{1.81} && & 28.41 & \textbf{55.49} & 1.03 \\ \midrule
S3D & Backbone~\cite{xie2018rethinking} && & 76.15 & 93.22 & - && & 27.69 & 54.68 & - \\
 & Non-local~\cite{wang2018non} && & 77.56 & 93.68 & 1.41 && & \textbf{29.52} & \textbf{56.91} & \textbf{1.83} \\
 & Ours - GPB && & \textbf{78.28} & \textbf{94.04} & \textbf{2.13} && & 29.23 & 56.22 & 1.54 
\\\bottomrule
\end{tabular}

\centering
\caption{Search results on Kinetics-600 and MiT using the differentiable method. Our attention cells consistently outperform non-local blocks on all the combinations of backbones and datasets. Position-specific attention cells (`Pos-Specific') consistently outperform position-agnostic attention cells (`Pos-Agnostic').}
\label{table-differentiable}
\begin{tabular}{llccccccccccc}
\toprule
 & && & \multicolumn{3}{c}{\textbf{Kinetics}} && & \multicolumn{3}{c}{\textbf{MiT}} \\  & Model && & Top-1 & Top-5 & $\Delta$Top-1 && & Top-1 & Top-5 & $\Delta$Top-1 \\ \midrule

I3D & Backbone~\cite{carreira2017quo} && & 75.58 & 92.93 & - && & 27.38 & 54.29 & - \\
 & Non-local~\cite{wang2018non} && & 76.87 & 93.44 & 1.29 && & 28.54 & 55.35 & 1.16 \\
 & Ours - Pos-Agnostic && & 77.56 & 93.63 & 1.98 && & 28.18 & 55.01 & 0.80 \\
 & Ours - Pos-Specific && & \textbf{77.86} & \textbf{93.75} & \textbf{2.28} && & \textbf{29.58} & \textbf{56.62} & \textbf{2.20} \\ \midrule
S3D & Backbone~\cite{xie2018rethinking} && & 76.15 & 93.22 & - && & 27.69 & 54.68 & - \\
 & Non-local~\cite{wang2018non} && & 77.56 & 93.68 & 1.41 && & 29.52 & 56.91 & 1.83 \\
 & Ours - Pos-Agnostic && & 77.82 & 93.72 & 1.67 && & 29.19 & 55.96 & 1.50 \\
 & Ours - Pos-Specific && & \textbf{78.51} & \textbf{93.88} & \textbf{2.36} && & \textbf{29.82} & \textbf{57.02} & \textbf{2.13}
 
\\\bottomrule
\end{tabular}

\end{table}

\begin{table}[t]
\centering
\caption{Generalization across different modalities (RGB to Optical flow).}
\label{table-flow}
\begin{tabular}{llccccccccccc}
\toprule
 & && & \multicolumn{3}{c}{\textbf{Kinetics}} && & \multicolumn{3}{c}{\textbf{MiT}} \\  & Model && & Top-1 & Top-5 & $\Delta$Top-1 && & Top-1 & Top-5 & $\Delta$Top-1 \\ \midrule

I3D & Backbone~\cite{carreira2017quo} && & 61.14 & 82.77 & - && & 20.01 & 42.42 & - \\
 & Non-local~\cite{wang2018non} && & 64.88 & 85.77 & 3.74 && & 21.86 & \textbf{46.59} & 1.85 \\
 & Ours - GPB && & 65.81 & 87.04 & 4.67 && & 21.83 & 45.45 & 1.82 \\
 & Ours - Differentiable && & \textbf{66.81} & \textbf{87.85} & \textbf{5.67} && & \textbf{21.94} & 45.57 & \textbf{1.93} \\ \midrule
S3D & Backbone~\cite{xie2018rethinking} && & 62.46 & 84.59 & - && & 20.50 & 42.86 & \textbf{-} \\
 & Non-local~\cite{wang2018non} && & 65.79 & 86.85 & 3.33 && & 22.13 & \textbf{46.48} & 1.63 \\
 & Ours - GPB && & \textbf{67.02} & \textbf{87.72} & \textbf{4.56} && & 22.29 & 46.16 & 1.79 \\
 & Ours - Differentiable && & 66.29 & 86.97 & 3.83 && & \textbf{22.52} & 46.30 & \textbf{2.02} 
 
\\\bottomrule
\end{tabular}
\end{table}

\begin{table}[t]
\centering
\caption{Generalization across different backbones.}
\label{table-backbone}
\begin{tabular}{llccccccccccc}
\toprule
 & && & \multicolumn{3}{c}{\textbf{Kinetics}} && & \multicolumn{3}{c}{\textbf{MiT}} \\  & Model && & Top-1 & Top-5 & $\Delta$Top-1 && & Top-1 & Top-5 & $\Delta$Top-1 \\ \midrule

I3D & Backbone~\cite{carreira2017quo} && & 75.58 & 92.93 & - && & 27.38 & 54.29 & - \\
 & S3D - GPB && & 77.47 & 93.67 & 1.89 && & 28.92 & 56.09 & 1.54 \\
 & S3D - Differentiable && & \textbf{77.81} & \textbf{93.74} & \textbf{2.23} && & \textbf{29.26} & \textbf{56.61} & \textbf{1.88} \\ \midrule
S3D & Backbone~\cite{xie2018rethinking} && & 76.15 & 93.22 & - && & 27.69 & 54.68 & - \\
 & I3D - GPB && & 78.23 & 94.07 & 2.08 && & 29.45 & 56.50 & 1.76 \\
 & I3D - Differentiable && & \textbf{78.46} & \textbf{94.05} & \textbf{2.31} && & \textbf{29.67} & \textbf{57.05} & \textbf{1.98} \\ \midrule
I3D-R50 & Backbone~\cite{wang2018non} && & 78.10 & 93.79 & - && & 30.63 & 58.15 & - \\
 & I3D - Differentiable && & \textbf{79.83} & \textbf{94.37} & \textbf{1.73} && & \textbf{32.48} & \textbf{60.31} & \textbf{1.85} \\
 & S3D - Differentiable && & 79.71 & 94.28 & 1.61 && & 31.91 & 59.87 & 1.28
\\\bottomrule
\end{tabular}
\end{table}

\begin{table}[t]
\centering
\caption{Generalization across different datasets.}
\label{table-dataset}
\begin{tabular}{llccccccccccc}
\toprule
 & && & \multicolumn{3}{c}{\textbf{MiT to Kinetics}} && & \multicolumn{3}{c}{\textbf{Kinetics to MiT}} \\  & Model && & Top-1 & Top-5 & $\Delta$Top-1 && & Top-1 & Top-5 & $\Delta$Top-1 \\ \midrule

I3D & Backbone~\cite{carreira2017quo} && & 75.58 & 92.93 & - && & 27.38 & 54.29 & - \\
 & GPB && & 77.34 & 93.47 & 1.76 && & 27.62 & 56.70 & 0.24 \\
 & Differentiable && & \textbf{77.85} & \textbf{93.89} & \textbf{2.27} && & \textbf{29.45} & \textbf{56.83} & \textbf{2.07} \\ \midrule
S3D & Backbone~\cite{xie2018rethinking} && & 76.15 & 93.22 & - && & 27.69 & 54.68 & - \\
 & GPB && & 77.54  & 93.62 & 1.39 && & 28.80 & 56.16 & 1.11 \\
 & Differentiable && & \textbf{78.19} & \textbf{93.98} & \textbf{2.04} && & \textbf{29.33} & \textbf{56.33} & \textbf{1.64}

\\\bottomrule
\end{tabular}

\end{table}

\begin{table}[t]
\caption{Comparison with the state-of-the-art methods. Our method (`I3D-R50+Cell') obtains similar or higher performance with the state-of-the-art methods on both Kinetics-600 and MiT.}
\label{table-sota}

\centering
\begin{subtable}[t]{0.47\textwidth}
\caption{Kinetics-600.}
\label{table-sota-k600}
\centering
\begin{tabular}{lccc}
\toprule

Model & Top-1 & Top-5 & GFLOPs  \\ \midrule
I3D~\cite{carreira2017quo} & 75.58 & 92.93 & 1136  \\
S3D~\cite{xie2018rethinking} & 76.15 & 93.22 & 656 \\
I3D-R50~\cite{wang2018non} & 78.10 & 93.79 & 938\\
D3D~\cite{stroud2020d3d} & 77.90 & - & - \\
I3D+NL~\cite{wang2018non} & 76.87 & 93.44 & 1305  \\
S3D+NL~\cite{wang2018non} & 77.56 & 93.68 & 825 \\
TSN-IRv2~\cite{wang2016temporal} & 76.22 &  - & 411  \\
StNet-IRv2~\cite{he2019stnet} & 78.99 & - & 440 \\
SlowFast-R50~\cite{feichtenhofer2019slowfast} & \textbf{79.9} & \textbf{94.5} & 1971 \\
\textbf{I3D-R50+Cell} & 79.83 & 94.37 & 1034
\\\bottomrule
\end{tabular}
\end{subtable}
\begin{subtable}[t]{0.52\textwidth}
\label{table-sota-mit}
\caption{MiT.}
\centering
\begin{tabular}{lccc}
\toprule

Model & Top-1 & Top-5 & Modality \\ \midrule

I3D~\cite{carreira2017quo} & 27.38 & 54.29 & RGB \\
S3D~\cite{xie2018rethinking} & 27.69 & 54.68 & RGB \\
I3D+NL~\cite{wang2018non} & 28.54 & 55.35 & RGB \\
S3D+NL~\cite{wang2018non} & 29.52 & 56.91 & RGB \\
R50-ImageNet~\cite{monfort2019moments} & 27.16 & 51.68 & RGB \\
TSN-Spatial~\cite{wang2016temporal} & 24.11 & 49.10 & RGB \\
I3D-R50~\cite{wang2018non} & 30.63 & 58.15 & RGB \\
\textbf{I3D-R50+Cell} & \textbf{32.48} & \textbf{60.31} & RGB \\\midrule
TSN-2stream~\cite{wang2016temporal} & 25.32 & 50.10 & R+F \\
TRN-Multiscale~\cite{zhou2018temporal} & 28.27 & 53.87 & R+F \\
AssembleNet-50~\cite{Ryoo2020AssembleNet:} & 31.41 & 58.33 & R+F 
\\\bottomrule
\end{tabular}
\end{subtable}

\end{table}

\subsection{Search Results}

Table~\ref{table-gpb} shows the search results of GPB and Table~\ref{table-differentiable} summarizes the search results using the differentiable method. Notably, attention cells found by the differentiable method can improve the accuracy of both backbones by more than 2\% on both datasets, and consistently outperform non-local blocks on all the combinations of backbones and datasets. 

In Table~\ref{table-differentiable}, `Pos-Agnostic' refers to that one attention design is learned for all the positions where the cells are inserted. `Pos-Specific' means that we learn a separate attention cell design for each position where a cell is inserted, i.e., the cells inserted at different positions can be different. We observe that position-specific attention cells consistently outperform position-agnostic attention cells.

\subsection{Generalization of Discovered Cells}
We examine how well the discovered attention cells can generalize to new settings. We do not perform any search in the following experiments, but directly apply attention cells searched for one setting to a different setting and see if the attention cells can improve the classification performance. Concretely, we evaluate whether our discovered attentions can generalize across different modalities, different backbones, and different datasets.

\textbf{Modality.} We insert the attention cells discovered on RGB frames into the backbone and train the network on optical flow only. The results are summarized in Table~\ref{table-flow}.  `GPB' refers to cells discovered by GPB and `Differentiable' refers to cells discovered by the differentiable method. Our attention cells significantly improve the classification accuracy when being applied on optical flow and consistently outperform non-local blocks for both backbones and on both datasets. For example, our attention cells improve the accuracy of I3D by 5.67\% on Kinetics-600. Note that the cells are discovered by maximizing its performance on RGB frames and no optical flow is involved during search. This demonstrates that our cells discovered on RGB frames can generalize well to optical flow.

\textbf{Backbone.} Table~\ref{table-backbone} summarizes the results of inserting cells discovered for one backbone to another backbone. The second row shows that cells discovered for S3D can still improve the classification accuracy of I3D by about 2\% on both datasets, even though these cells are never optimized to improve the performance of I3D.
We observe similar improvement when inserting cells found for I3D to S3D (third row), or cells found for I3D/S3D to I3D-R50 (last row). Notably, our attention cells can still outperform non-local blocks even after being inserted into a different backbone. For example, cells found for S3D achieve 77.81\% accuracy on Kinetics-600 after being inserted to I3D, which outperforms non-local blocks (76.87\%) and performs similar to cells specifically discovered for I3D (77.86\%).

\textbf{Dataset.} We insert attention cells discovered on MiT to the corresponding backbone, fully train the network on Kinetics-600 and report its accuracy on Kinetics-600 in the middle column (`MiT to Kinetics') of Table~\ref{table-dataset}. We observe that cells discovered on MiT can improve the accuracy on Kinetics-600 by more than 2\%, although they are never optimized to improve the Kinetics-600 performance during the search. Similarly, the right column (`Kinetics to MiT') demonstrates that the cells searched on Kinetics-600 can also generalize gracefully to MiT. We conclude that our attention cells generalize well across datasets.

\subsection{Comparison with State-of-the-art}
We insert our attention cells found on I3D into I3D-R50 (`I3D-R50+Cell') and compare with the state-of-the-art methods in Table~\ref{table-sota}. On Kinetics-600, we obtain similar performance with SlowFast-R50~\cite{feichtenhofer2019slowfast} with fewer inference FLOPs. On MiT, we achieve 32.48\% top-1 accuracy and 60.31\% top-5 accuracy only using the RGB frames. This significantly outperforms the previous state-of-the-art method AssembleNet-50~\cite{Ryoo2020AssembleNet:}, which uses both RGB frames and optical flow.  

\section{Conclusions}
We propose a novel search space for spatiotemporal attention cells for the application of video classification. We also propose a differentiable formulation of the search space, allowing us to learn position-specific attention cell designs with zero extra cost compared to learning a single position-agnostic attention cell. We show the significance of our discovered attention cells on two large-scale video classifications benchmarks. The discovered attention cells also outperform non-local blocks and demonstrate strong generalization performance when being applied to different modalities, backbones, or datasets.

\bigbreak
\noindent\textbf{Acknowledgement.} We thank Guanhang Wu and Yinxiao Li for insightful discussions and the larger Google Cloud Video AI team for the support.

%
%
\bibliographystyle{splncs04}
\bibliography{egbib}

\setcounter{section}{0}
\setcounter{table}{0}
\setcounter{figure}{0}
\setcounter{equation}{0}
\renewcommand\thesection{\Alph{section}}
\renewcommand\thetable{\Alph{table}}
\renewcommand\thefigure{\Alph{figure}}
\renewcommand\theequation{\Alph{equation}}

\title{AttentionNAS: Spatiotemporal Attention Cell Search for Video Classification \\ Supplementary Materials}
\titlerunning{AttentionNAS}
\author{}
\institute{}
\maketitle

\section{Attention Cell Search Space Details}
\subsection{Keys and Values in Dot-product Attention}
We introduce an additional design choice in dot-product attention. In Sec 3.1, a dot-product attention operation is defined as:
\begin{equation}
\label{eq-dotproductatt-def1}
\begin{split}
f'_\text{in} &= \texttt{ReshapeTo2D}(f_\text{in}), \\
W_\text{dot-prod} & = \phi(G_1(f'_\text{in})  G_2(f'_\text{in})^{T}), \\
f_\text{out} & = \texttt{ReshapeTo2D}^{-1} (  W \texttt{ReshapeTo2D}(G_3(f_\text{in})) ),
\end{split}
\end{equation}
where $f_\text{in}$ and $f_\text{out}$ are the input and output feature map of the attention operation respectively, $W_\text{dot-prod}$ is the attention weight matrix, and $G_1$, $G_2$ and $G_3$ are all $1\times 1 \times 1$ convolutional layers.

Let $Q = G_1(f'_\text{in})$, $K = G_2(f'_\text{in})$ and $V=\texttt{ReshapeTo2D}(G_3(f_\text{in}))$. $Q$, $K$ and $V$ are termed as query, keys and values in dot-product attention~\cite{vaswani2017attention}. In Eq~\ref{eq-dotproductatt-def1}, the query, keys and values are computed based on the same feature map, i.e., the operation input $f_\text{in}$. It is also common practice in dot-product attention to compute the keys and values based on feature maps other than $f_\text{in}$. For example, dot-product attention has been used in Transformer~\cite{vaswani2017attention} in the following way: the query comes from the decoder while the keys and values come from the encoder, so that every position in the decoded sequence can attend to positions in the input sequence.

In our search space, for a dot-product attention operationin, we also allow computing its keys and values based on the cell input $f_0$. This allows positions in the operation input $f_\text{in}$ to attend to positions in the cell input $f_0$. When computing keys and values based on $f_0$, the dot-product attention becomes:
\begin{equation}
\label{eq-dotproductatt-def2}
\begin{split}
f'_0 &= \texttt{ReshapeTo2D}(f_0), \\
W_\text{dot-prod} & = \phi(G_1(f'_\text{in})  G_2(\bm{\textcolor{red}{f'_{0}}})^{T}), \\
f_\text{out} & = \texttt{ReshapeTo2D}^{-1} (  W \texttt{ReshapeTo2D}(G_3(\bm{\textcolor{red}{f_0}})) ).
\end{split}
\end{equation}
The differences between Eq.~\ref{eq-dotproductatt-def1} and Eq.~\ref{eq-dotproductatt-def2} are highlighted in boldface and red.

In summary, for dot-product attention operations in the attention cell, we can choose to compute its keys and values based on the operation input $f_\text{in}$ or the cell input $f_0$. We include this choice in the cell level search space, i.e., all the dot-product attention operations make the same choice, either computing the keys and values based on their own operation input or the cell input $f_0$.

\subsection{Channel Attention}
While our search space mainly focuses on spatiotemporal attention, we include channel attention as an additional choice in the search space. Concretely, when building an attention operation, the search algorithm can choose whether to apply a feature gating layer~\cite{xie2018rethinking} to the attended feature map $f_\text{out}$. The feature gating layer is a typical channel attention mechanism. It first applies average pooling to a 4D feature map across space and time, then learns a weighting factor for each channel, and finally multiplies features at each channel of the original feature map with the learned weighting factor. Note that channel attention does not replace the attention operation described above and is only an additional layer choice within the attention operation.

When using differentiable search, we learn a 2-dim probability distribution $w_{i, j}^{\text{gating}}$ for each node, indicating whether to include a feature gating layer~\cite{xie2018rethinking} in the attention operation represented by the node.

\section{Experimental Details}

\subsection{Training and Inference}

We conduct experiments on two benchmark datasets: Kinetics-600~\cite{carreira2018short} and Moments in Time (MiT)~\cite{monfort2019moments}. Kinetics-600 contains about 392K training videos and 30K validation videos from 600 classes. MiT consists of about 800K training videos and 34K validation videos from 339 classes.

After obtaining the attention cells found by our method, we fully train the backbone networks and cells on training videos and report their performance on validation videos. Following non-local blocks~\cite{wang2018non}, we insert 5 cells or non-local blocks into the backbone. For I3D or S3D, they are inserted 5 inception modules (4a to 4e, see Table 1 in~\cite{szegedy2015going}). For I3D-R50, we insert them after 5 residual blocks, where 2 cells are inserted after every other residual block in res$_3$ and 3 cells are inserted after every other residual block in res$_4$.

During training, we resize the spatial resolution of videos to $256\times 256$ and randomly crop $224\times 224$ pixels or its horizontal flip from videos, for both Kinetics-600~\cite{carreira2018short} and MiT~\cite{monfort2019moments}. For I3D or S3D, we randomly crop 64 consecutive frames from the full-length video as the input clip during training. For I3D-R50, we randomly crop 16 frames with stride 4 from the full-length video.

During inference, we perform fully-convolutional inference both spatially and temporally. We resize the spatial resolution to $256\times 256$, pass the full-length video to the network, and predict the class based on the softmax scores. Our inference procedure does not require the sampling of multiple temporal clips and spatial crops in previous works~\cite{feichtenhofer2019slowfast}. The input clip to I3D or I3D has 250 frames for Kinetics-600 and has 75 frames for MiT. The input clip to I3D-50 has 64 frames for Kinetics-600 and has 18 frames for MiT, which is obtained by temporally downsampling the full-length video with stride 4.

We initialize the backbone in all the models (backbone, backbone + non-local blocks or our attention cells) with its ImageNet pre-trained weights. I3D or S3D based models are trained for 135 epochs, and I3D-R50 based models are trained for 150 epochs on Kinetics-600. All the models are trained for 45 epochs on MiT. We adopt a cosine learning rate schedule with a linear warm-up. The initial learning rate is 0.1 for I3D or S3D and 0.4 for I3D-R50. All the models are trained on 50 GPUs with synchronized SGD. The momentum is 0.9. The batch size per GPU is 6 for I3D or S3D and 4 for I3D-R50.

\subsection{Attention Cell Implementation}

We have three pre-processing steps for the input to the entire attention cell: (1) channel reduction, (2) spatial resize, and (3) temporal grouping. These steps can not only reduce the computation consumed by the cell, but also allow the cell to process feature maps of different temporal and spatial resolutions.

Let $(B, T, H, W, C)$ be the shape of the input to the entire cell. We explicitly write out the batch size dimension $B$ for better explanation. We first reduce the number of channels from $C$ to $C_\text{reduction}$ with a $1 \times 1 \times 1$  convolutional layer. After channel reduction, the shape becomes $(B, T, H, W, C_\text{reduction})$. Then, we resize the spatial resolution of the feature map with bilinear interpolation from $(H, W)$ to $(H_\text{resize}, W_\text{resize})$, so the shape becomes $(B, T, H_\text{resize}, W_\text{resize}, C_\text{reduction})$. Finally, we divide the feature map into multiple groups of $T_\text{group}$ frames and obtain a feature map of shape $(nB, T_\text{group}, H_\text{resize}, W_\text{resize}, C_\text{reduction})$, where $T = n \cdot T_\text{group}$ and zero padding frames are added when necessary. The feature map of shape $(nB, T_\text{group}, H_\text{resize}, W_\text{resize}, C_\text{reduction})$ is then passed to attention operations in the cell. During the combination procedure, we resize the spatial resolution back to $(H, W)$ and merge temporal groups back to $T$ frames.

It is not difficult to see that these steps can reduce the computation. We elaborate on the second advantage. Note that the temporal and spatial resolution of test videos can vary (e.g., $250 \times 256 \times 256$) and be different from sampled training clips (e.g., $64 \times 224 \times 224$). This causes
the shape of the feature map output by each layer to be different between training and test. However, temporal attention requires the spatial resolution of the feature map to be fixed and spatial attention requires the number of frames to be fixed. To address this issue, we adopt these pre-processing steps so that the input to attention operations always has a fixed shape of $(T_\text{group}, H_\text{resize}, W_\text{resize}, C_\text{reduction})$.

\subsection{Search Algorithm}

\subsubsection{GPB}
We sample training videos from the original dataset as the search-train and search-validation split. No validation videos are used during the search. We maximize the validation performance using GPB. We set the number of trails of GPB to 50, i.e., 50 attention cells are sampled by GPB and evaluated on the search-validation split after trained on the search-train split. Both the search-train split for Kinetics-600 and MiT contain about 360K videos. We train for 60 epochs for Kinetics-600 and 20 epochs for MiT during the search on their corresponding search-train split. We set $K=4$ and search for an attention cell consisting of 4 attention operations. We use GPB to find one position-agnostic attention cell and insert the same cell architecture at different positions in the backbone network. To simplify the search space explored by GPB, we restrict the $k^{th}$ operation to select only one feature from $\{f_0, f_1, \ldots, f_{k-1}\}$ as its input.

\subsubsection{Differentiable Method}
\label{sec-expdetail-search-algo-diff}
When using the differentiable search method, we consider a supergraph consisting of 2 levels. Each level in the supergraph has 6 nodes. We do not include more nodes in one level due to the GPU memory constraint. At eacl level, we repeat each attention dimension twice and only include dot-product attention. So the 6 nodes are 2 temporal dot-product, 2 spatial dot-product, and 2 spatiotemporal dot-product attention operations. We also fix that the keys and values of dot-product attention are computed based on the attention cell input (see Eq.~\ref{eq-dotproductatt-def2}). This is the default supergraph design and we study other supergraph designs in Sec~\ref{sec-ablation-supergraph}.

The connection weights and the network weights are learned jointly on training videos. The entire search process of the differentiable method consumes a computational cost similar to fully training one network on the training videos. For example, training I3D with the found attention cells on Kinetics-600 takes about 2.5 days. Searching attention cells for I3D, i.e., training I3D with supergraphs, takes about 3.5 days on Kinetics-600. The increase in the time is due to that supergraphs consume more computation than the final attention cells.

When deriving the attention cell design from the learned connection weights, the hyper-parameters $\alpha$ and $\beta$ are set to $\alpha=3, \beta=2$. Attention cells found by the differentiable method do not have a fixed number of operations, which are determined by the determined connection weights and $\alpha$ and $\beta$. Each operation may receive up to $\beta$ feature maps and computes a weighted sum of these feature maps as its input. We slightly revisit the combination procedure for cells found by the differentiable method. Instead of combining all the operation output feature maps, we only combine the output of the top $\alpha$ nodes (operations) with the highest weights in $w^{\text{sink}}$.

\subsection{Comparison of FLOPs}

\begin{table}[!htbp]
\centering
\caption{Inference FLOPs on Kinetics-600.}
\label{table-flops}
\begin{tabular}{lcccccc}
\toprule
Model && & Top-1 & Top-5 &  GFLOPs \\ \midrule
I3D~\cite{carreira2017quo} && & 75.58 & 92.93 & 1136  \\
I3D+NL~\cite{wang2018non} && & 76.87 & 93.44 & 1305  \\
\textbf{I3D+Cell} &&  & 77.86 & 93.75 & 1170 \\
\midrule
S3D~\cite{xie2018rethinking} && & 76.15 & 93.22 & 656 \\
S3D+NL~\cite{wang2018non} && & 77.56 & 93.68 & 825 \\
\textbf{S3D+Cell} && & 78.51 & 93.88 & 692 \\
\midrule
I3D-R50~\cite{wang2018non} && & 78.10 & 93.79 & 938\\
\textbf{I3D-R50+Cell} && & 79.83 & 94.37 & 1034
\\\bottomrule
\end{tabular}
\end{table}

We compare the inference FLOPs of all the models on Kinetics-600 in Table~\ref{table-flops}. Note that although our cells contain multiple operations, the aforementioned pre-processing applied on the cell input can effectively reduce the FLOPs consumed by attention operations. As shown in Table~\ref{table-flops}, our cells only add a small amount of computation to the backbone network and use fewer FLOPs than non-local blocks. The FLOPs are computed when the input clip has 250 frames with spatial resolution $256 \times 256$.

\section{Ablation Study of Supergraph Designs}
\label{sec-ablation-supergraph}

In the differentiable method, we represent the attention cell search space as a supergraph. Using different supergraph designs allows us to analyze what desgin choice is important for the performance of the discovered attention cells. Specifically, we compare the following three supergraph designs:

\begin{table}[!htbp]
\centering
\caption{Comparison between different supergraph designs.}
\label{table-supergraph-ablation}
\begin{tabular}{lcccccc}
\toprule
Model && & Top-1 & Top-5  \\ \midrule
I3D~\cite{carreira2017quo} && & 75.58 & 92.93  \\
I3D+NL~\cite{wang2018non} && & 76.87 & 93.44 \\
I3D+SG-1 Cell && & 77.86 & 93.75 \\
I3D+SG-2 Cell && & 77.82 & 93.75 \\
I3D+SG-3 Cell && & 77.71 & 93.87
\\\bottomrule
\end{tabular}
\end{table}

\textbf{SG-1.} SG-1 is our default choice described Sec~\ref{sec-expdetail-search-algo-diff}. It contains 2 levels, where each level has 6 nodes. SG-1 only contains dot-product attention and the 6 nodes at each level are 2 temporal dot-product, 2 spatial dot-product, and 2 spatiotemporal dot-product attention operations. In SG-1, the keys and values of dot-product attention are computed based on the cell input (see Eq.~\ref{eq-dotproductatt-def2}).

\textbf{SG-2.} Same SG-1, SG-2 also contains 2 levels and each level has 6 nodes. SG-2 include both map-based attention and dot-product attention. The 6 nodes at each level are 1 temporal dot-product, 1 spatial dot-product, 1 spatiotemporal dot-product, 1 temporal map-based, 1 spatial map-based, and 1 spatiotemporal map-based attention operation. In SG-2, the keys and values of dot-product attention are also computed based on the cell input (see Eq.~\ref{eq-dotproductatt-def2}).

\textbf{SG-3.} SG-3 is the same as SG-1 except that the keys and values of dot-product attention are computed based on the input to each attention operation (see Eq.~\ref{eq-dotproductatt-def1}), instead of the cell input.

Comparing SG-1 and SG-2 tells us which attention type (map-based or dot-product) is more important. As shown in Table~\ref{table-supergraph-ablation}, SG-1 and SG-2 achieve a very close top-1 accuracy and the same top-5 accuracy on Kinetics-600. However, we observe that most operations (20 of out 28) in the 5 position-specific cells discovered from SG-2 are dot-product attention. This shows that dot-product attention is more important than map-based attention, and explains why SG-1 can achieve high accuracy with only dot-product attention.

SG-3 achieves similar performance with SG-1 and also outperforms non-local blocks. This shows that our search space is not sensitive to whether to compute the keys and values based on the input to each dot-product operation or based on the cell input.

\section{Attention Cell Visualization}

\begin{figure}[!htbp]
\centering
\includegraphics[width=0.85\columnwidth]{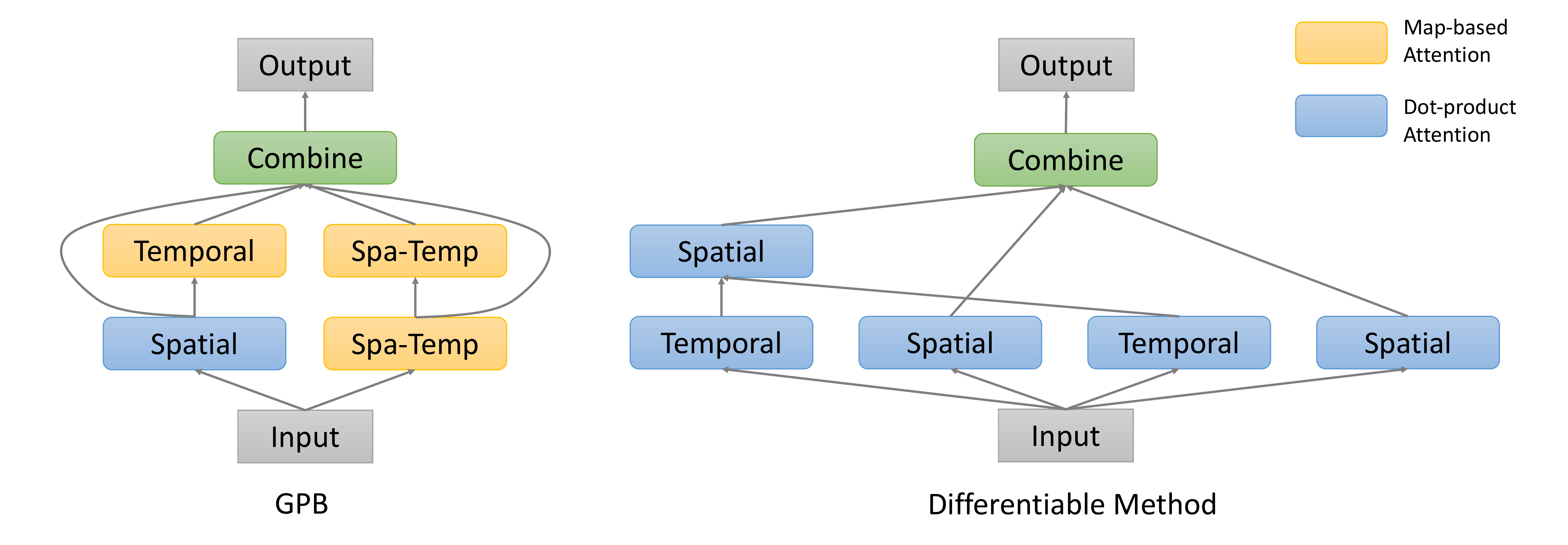}
\caption{Visualization of the position-agnostic cell discovered by GPB and the differentiable method for I3D and on Kinetics-600. `Spa-Temp' stands for the spatiotemporal attention dimension.}
\label{fig-cell-pos-agnostic}

\centering
\includegraphics[width=0.85\columnwidth]{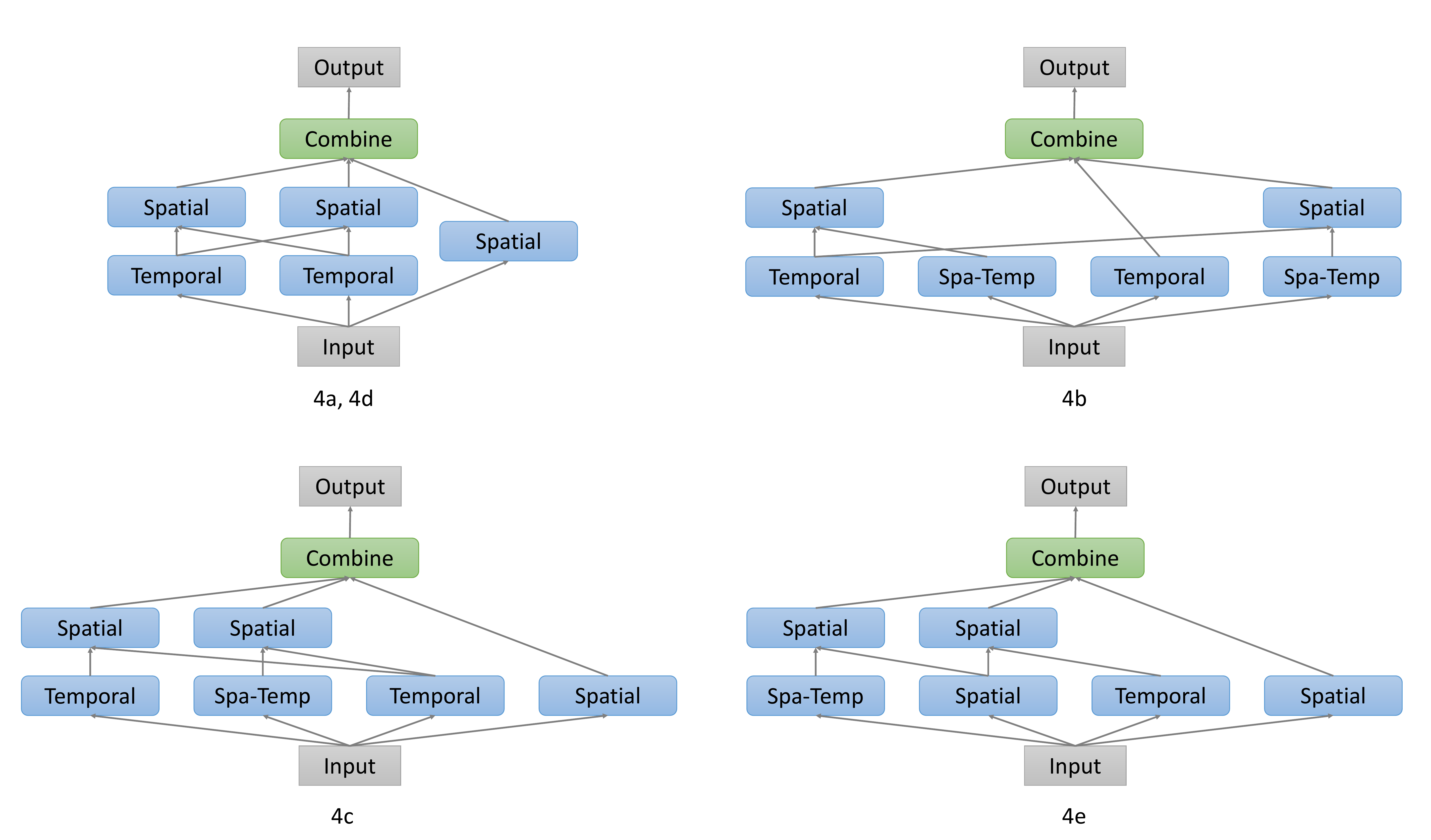}
\caption{Visualization of the position-specific cells discovered by the differentiable method for I3D and on Kinetics-600. `Spa-Temp' stands for the spatiotemporal attention dimension. The text under each cell indicates the inception module after which the cell is inserted (4a to 4e, see Table 1 in \cite{szegedy2015going}) in the Inception network. The learned attention cell for 4a and 4d are the same.}
\label{fig-cell-pos-specific}
\end{figure}
We visualize the position-agnostic attention cell found by GPB and the differentiable method in Fig.~\ref{fig-cell-pos-agnostic}. The position-specific cells found by the differentiable method are shown in Fig.~\ref{fig-cell-pos-specific}. These cells are found for I3D and on Kinetics-600. We show the attention dimension and type of each operation, as well as the connectivity between the operations.

The cell found by GPB contains both map-based attention and dot-product attention and contains one path that first applies spatial attention and then temporal attention. Cells found by the differentiable method only contain dot-product attention as we only include dot-product attention in the supergraph (SG-1). We observe that all the cells found by the differentiable method  prefer decomposing spatiotemporal attention into temporal and spatial attention, as they all contain paths that first apply temporal attention and then spatial attention. This shares a similar spirit to S3D~\cite{xie2018rethinking} that decomposes a 3D convolution into a 2D spatial convolution and a 1D temporal convolution. As a side note, our cells choose to first apply temporal and then spatial attention, while S3D first applies spatial convolution and then temporal convolution.

\end{document}